\documentclass{egpubl}
\usepackage{eurovis2020}

\SpecialIssuePaper         

\usepackage[T1]{fontenc}
\usepackage{dfadobe}  

\usepackage{cite}  
\BibtexOrBiblatex
\electronicVersion
\PrintedOrElectronic
\ifpdf \usepackage[pdftex]{graphicx} \pdfcompresslevel=9
\else \usepackage[dvips]{graphicx} \fi

\usepackage{pifont}
\usepackage{egweblnk}

\newcommand{\thomas}{Expert~\#3}
\newcommand{\edward}{Expert~\#1}
\newcommand{\arthur}{Expert~\#2}

\newcommand{\Thomas}{Expert~\#3}
\newcommand{\Edward}{Expert~\#1}
\newcommand{\Arthur}{Expert~\#2}

\newcommand{\figref}[2]{~(Fig.~\ref{#1}~\ding{#2})}

\newcommand{\tool}{DRLViz}
\newcommand{\vizdoom}{ViZDoom}

\newcommand{\agent}{agent}

\usepackage{xcolor}

\usepackage{xspace}
\newcommand{\ie}{i.e.,\xspace}

\newcommand{\eg}{e.g.,\xspace}

\usepackage{enumitem}

\title[Understanding Decisions and Memory in Deep Reinforcement Learning]%
      {DRLViz: Understanding Decisions and Memory in Deep Reinforcement Learning}

\author[T. Jaunet, R. Vuillemot and C. Wolf]
{\parbox{\textwidth}{\centering  \vspace{-8.5mm}T. Jaunet$^{1}$
        \, R. Vuillemot$^{2}$ and C. Wolf$^{1,3}$ 
        }\vspace{-4.5mm}
        \\
{\parbox{\textwidth}{\centering $^1$ LIRIS, INSA-Lyon, France\\
         $^2$LIRIS, École Centrale-Lyon, France\\
         $^3$ CITI, INRIA, France
      }
}
}

\begin{document}

\teaser{
\vspace{-8.5mm}
 \includegraphics[width=15.5cm]{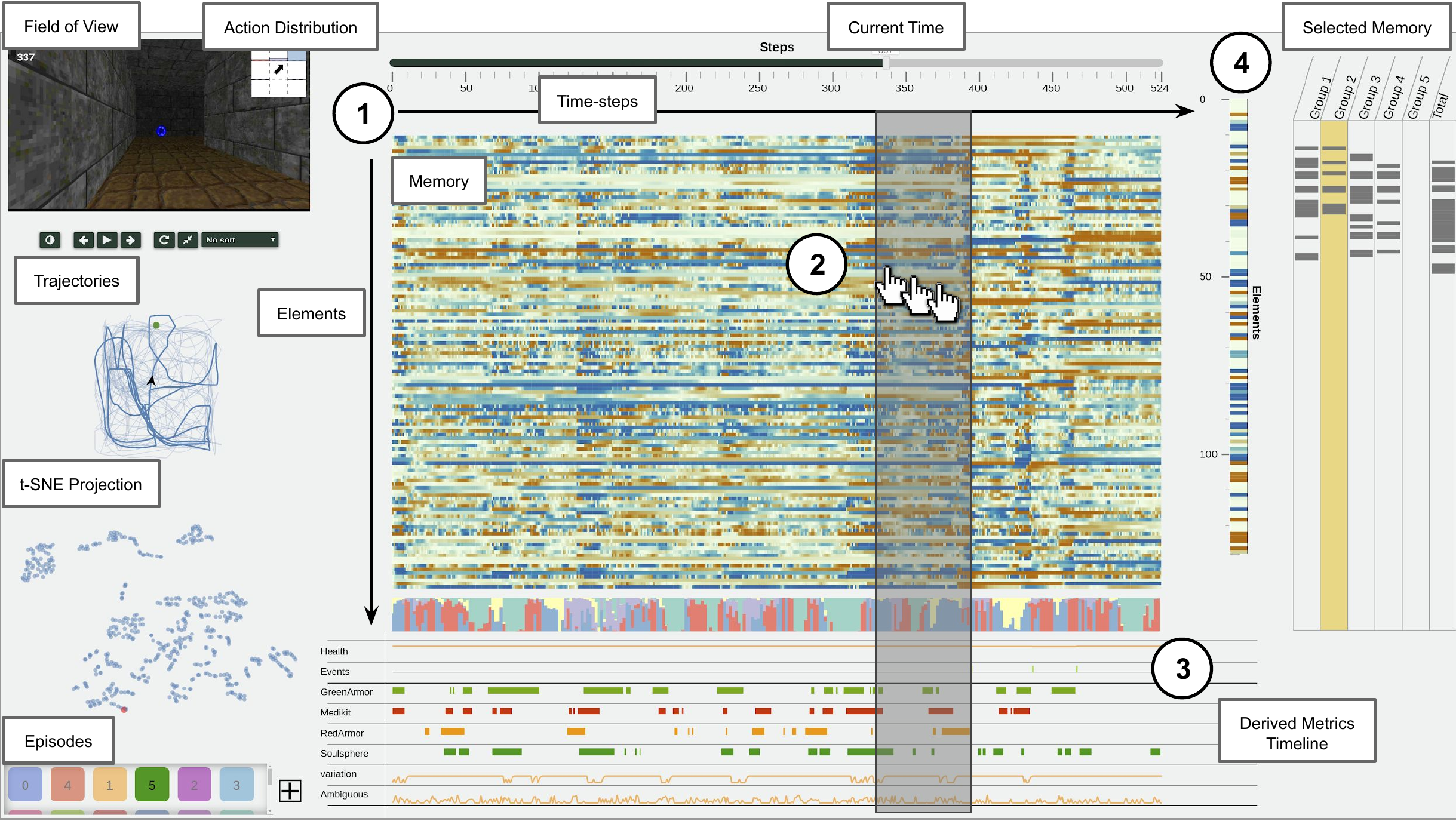}
 \centering
  \caption{ \tool\ displays a trained agent memory, which is a large temporal vector, as a horizontal heat-map~\ding{192}. Analysts can \emph{browse} this memory following its temporal construction; \emph{filter} according to movements of the agent and derived metrics we calculated~\ding{193} (\eg when an item is in the field of view~\ding{194}); and \emph{select the memory} to filter elements and compare them~\ding{195}.}
\label{fig:teaser}
}

\maketitle

\begin{abstract}
We present DRLViz, a visual analytics interface to interpret the internal memory of an agent (e.g. a robot) trained using deep reinforcement learning. This memory is composed of large temporal vectors updated when the agent moves in an environment and is not trivial to understand due to the number of dimensions, dependencies to past vectors, spatial/temporal correlations, and co-correlation between dimensions. It is often referred to as a black box as only inputs (images) and outputs (actions) are intelligible for humans. Using DRLViz, experts are assisted to interpret decisions using memory reduction interactions, and to investigate the role of parts of the memory when errors have been made (e.g. wrong direction). We report on DRLViz applied in the context of video games simulators (ViZDoom) for a navigation scenario with item gathering tasks. We also report on experts evaluation using DRLViz, and applicability of DRLViz to other scenarios and navigation problems beyond simulation games, as well as its contribution to black box models interpretability and explain-ability in the field of visual analytics.


\end{abstract}

\section{Introduction}
Automatic navigation is among the most challenging problems in Computer Science with a wide range of applications, from finding shortest paths between pairs of points, to efficiently exploring and covering unknown environments, up to complex semantic visual problems (``\emph{Where are my keys?}''). Addressing such problems is important for modern applications such as autonomous vehicles to improve urban mobility, social robots and assisting elderly people. Historically, navigation was often solved with discrete optimization algorithms such as Dijkstra~\cite{Dijkstra1959}, A-Star~\cite{AStar1968}, Front-propagation~\cite{YamauchiFrontiers1997} etc., applied in settings where spatial maps are constructed simultaneously with solving the navigation problem. These algorithms are well understood, but are restricted to simple waypoint navigation. Recently, techniques from Machine/Deep Learning have shown spectacular progress on more complex tasks involving visual recognition, and in particular in settings where the agent is required to discover the problem statement itself from data. In particular, Reinforcement Learning (RL) and the underlying Markov Decision Processes (MDP) provide a mathematically founded framework for a class of problems focusing on interactions between an \agent\ and its environment~\cite{Sutton2018ReinforcementIntroduction}. In combination with deep networks as function approximators, this kind of models was very successful in problems like game playing~\cite{mnih2015humanlevel,Silver2017MasteringKnowledge}, navigation in simulated environments~\cite{embodiedqa,gordon2018iqa,ParisottoNeuralMap2018}, and work in human-computer interaction (HCI) emerging\cite{DebardECMLPKDD2019}.

The goal of Deep Reinforcement Learning (DRL) is to train agents which interact with an environment. The agent sequentially takes decisions $a_t$, where $t$ is a time instant, and receives a scalar reward $R_t$, as well as a new observation $o_t$. The reward encodes the success of the agent's behavior, but a reward $R_t$ at time $t$ does not necessarily reflect the quality of the agent's action at time $t$. As an example, if an agent is to steer an autonomous vehicle, receiving a (very) negative reward at some instant because the car is crashed into a wall, this reflects a sequence of actions taking earlier then the last action right before the crash, which is known as the \emph{credit assignment problem}. The reinforcement learning process aims at learning an optimal policy of actions which optimizes the expected accumulated future reward $V_t = \sum_{t'=t}^{t+\tau} R_t$ over a horizon $\tau$.

If agents trained with DRL were deployed to real life scenarios, failures and unexpected behaviors~\cite{lehman2018surprising} could lead to severe consequences. This raises new concerns in understanding on what ground models' decisions (\eg brake) are based~\cite{ribeiro2016should}. To assess the decision of a trained model, developers~\cite{Hohman2019VisualFrontiers} must explore its context (\eg a pedestrian on the road, speed, previous decisions) and associate it with millions of deep networks parameters which is not feasible manually. Analysing a decision after-the-fact, referred to as post-hoc interpretability~\cite{Lipton2016TheInterpretability}, has been a common approach in visualization. It consists in collecting any relevant information such as inputs and inner-representations produced while the model outputs decision. With such an approach, DRL experts explore their models without having to modify them and face the trade-off between interpretability and performances. Visual analytics for post-hoc interpretability~\cite{Hohman2019VisualFrontiers} yields promising results on tasks such as image classification~\cite{olah2018the}, or text prediction~\cite{Strobelt2017LSTMVis:Networks}; however, it remains an under-explored challenge for DRL \emph{with} memory.

We built \tool, a novel Visual Analytics interface aimed at making Deep Reinforcement Learning models with memory more interpretable for experts. \tool\ exposes a trained agent's memory using a set of interactive visualizations the user can overview and filter, to identify sub-sets of the memory involved in the agent's decision. \tool\ targets expert users in DRL, who are used to work with such models (referred to as \emph{developers} in~\cite{Hohman2019VisualFrontiers}). Typically, those experts have already trained agents and want to investigate their decision-making process. We validated \tool\ usability with three experts and report on their findings that informed us on future improvement such as applicability to other scenarios, and novel interactions to reduce the memory of an agent and better find patterns within it.

\section{Context and Background} 

The context of our work is related to building deep neural network models to train robots achieving human assistance tasks in real-world environments. As the sample efficiency of current RL algorithms is limited, training requires a massive amount of interactions of the agent with the environment --- typically in the order of a billion. Simulators can provide this amount of interactions in a reasonable time frame, and enable to work with a constantly controlled world, that will generate less noise (\eg a shade) in the agent's latent representation. We will discuss in the perspectives section the extension of our work beyond simulators and the knowledge transfer from simulation to real-world scenarios, where variability (\eg lighting, clouds, shades, etc.) and non-deterministic behaviors (\eg robots may turn more or less depending on its battery charge) occur.

\subsection{Navigation Problem Definitions}
 
\begin{figure}[ht!]
    \centering
\includegraphics[width=7.5cm]{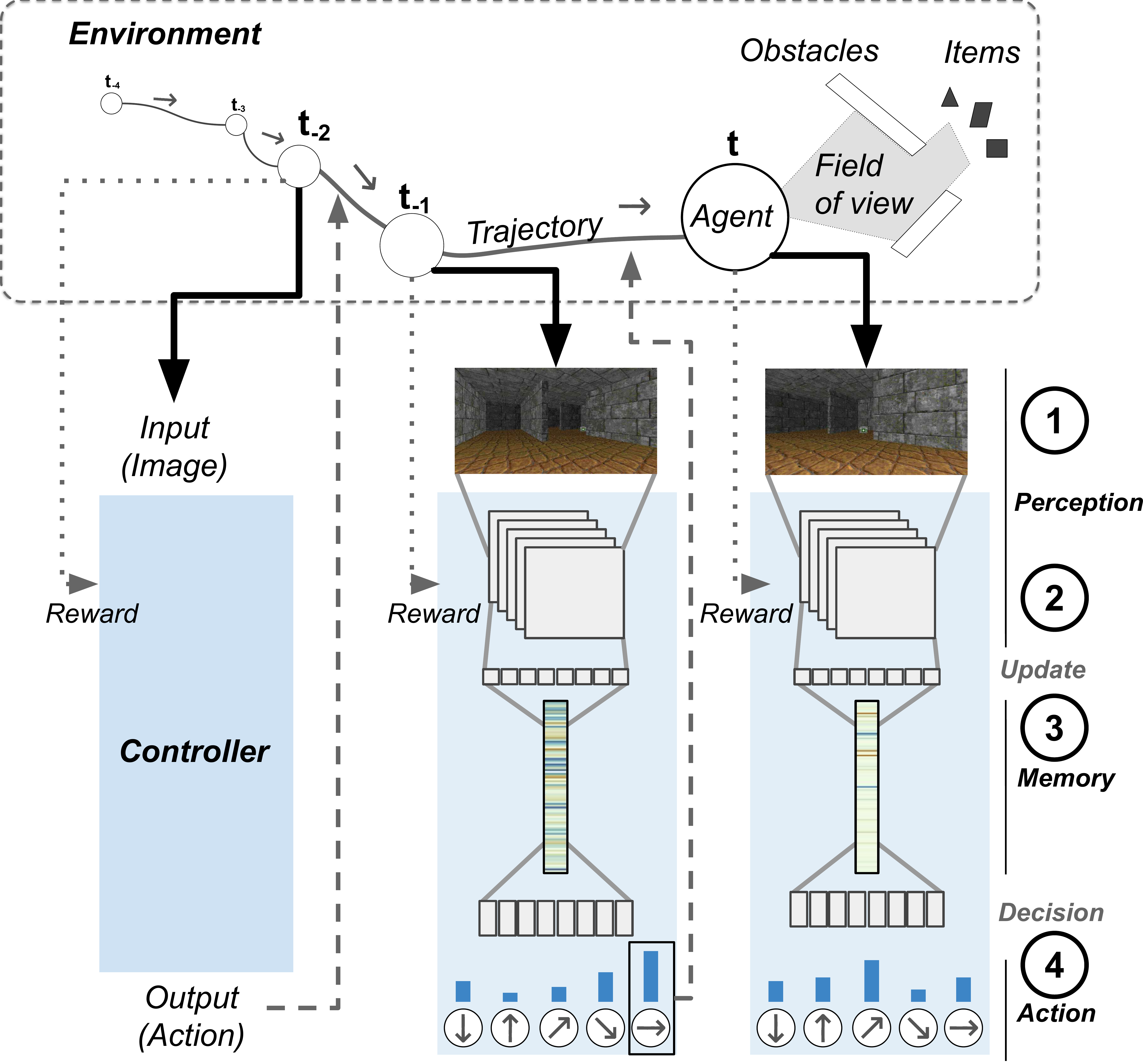}
    \caption{Our navigation problem consists in solving a visual task (\eg fetch, interact, or recognize items) while avoiding obstacles in an environment. Deep Reinforcement Learning can be used to solve this problem by using an image as input~\ding{172} at time $t$. Features are then extracted from this image~\ding{173}, and combined with the previous memory vector $t-1$ \ding{174}. Using this memory vector, the agent decides to move forward or turn left, for instance~\ding{175}.}
    \label{fig:overview}
    \vspace{-.5cm}
\end{figure}

Our focus is on navigation problems, where an \emph{agent} (\eg robot, human) moves within a 2D space we call \emph{environment} (Fig.~\ref{fig:overview}). 
An environment contains obstacles (\eg walls), items the agent may want to gather or avoid, and is usually bounded (\eg a room). The goal of the agent can vary according to the problem variation, but typically is to reach a particular location (\eg gather items, find a particular spot). Importantly, the goal itself needs to be discovered by the agent through feedback in the form of a scalar reward signal the environment provides: for instance, hitting a wall may provide negative reward, finding a certain item may result in positive reward. To discover and achieve the goal, the agent must explore its environment using actions. In our case, those actions are discrete and elements of the following alphabet: $a{\in}A$, with $A = $\emph{\{forward, forward+right, right, left, forward+left\}}. The navigation task ends when the agent reaches its goal, or when it fails (\eg dies, timeout).

As the agent explores its environment, it produces a trajectory. A trajectory is a series of positions $p$ ($x$,$y$) in a space $S$ bounded by the environment. Those positions are ordered by time-step $t \in T$, where $t_0 < t_1 < t_n$, and the interval between $t_n$ and $t_{n+1}$ is the time the agent takes to act. 
In addition to positions, trajectories contain complementary attributes $b$, which may vary depending on the agent goal (\eg number of gathered items, velocity, etc.). We call step$_t$ the combination of both the agent position $p$ and its attributes $b$, at a given time-step $t$. Thus $step_t$ can be represented as follows ${<}p_t,b_t{>}$.
The transition between steps occurs as the agent makes a decision. An \emph{episode} groups all iterations from the first step at $t_0$, until the agent wins or looses at $t_n$.

\subsection{Navigation using the \vizdoom\ Simulation}

The simulation environment we use to train agents to navigate is \vizdoom~\cite{Kempka2016ViZDoom:Learningb} which provides instances of the navigation problem based on Doom, a very popular video game in the 90's. \vizdoom\ is a 3D world and as such is a proxy problem to mobile service robotics. It supplies different scenarios focusing on various goals (\eg survive, reach a location, gather items, avoid enemies, etc.). For expert evaluation, and Fig.~\ref{fig:overview} we used the \emph{k-items} scenario from~\cite{BeechingDeepSupercomputer} with $k=4$. In this scenario, the agent, walls and items are randomly placed in an environment at the beginning of each episode. Then the agent needs to explore the environment until it succeed, fail or reach a timeout of $525$ steps. To succeed the agent must first gather a green armor, then a red armor, followed by a health pack, and finally a soul-sphere (blue circle). Gathering the items in another order instantly kills the agent and ends the episode (fail). Gathering an item in the right order grants a $+0.5$ reward $r$, while failing grants a reward of $-0.25$. Additionally, the agent receives a reward of $-0.0001$ at each step. Despite \vizdoom\ being a 3D world, the agent positions $p$ are within a bounded continuous 2D plane corresponding to the bird's eye view of the environment. We summarize a time-step $t$ as follows: ${<}p_t,(r_t){>}$. 

This task is challenging as the agent is required to take a decision on the next action based on partial information of the environment, \ie the task is partial observable. The observed image represents the agent's field of view (\ie what is in front of it), in a 90 degree range and unlimited depth. The agent is required to recall previously seen observations in some way as it doesn't have access to a global view of the map. These views are stored in its latent memory, the representation studied in this work. The agent should also use its memory to encode information on the items it gathered, and the positions of items or walls encountered in order to quickly complete this task.

\subsection{Deep Reinforcement Learning and Memory}
\label{sec:drl}

As expressed in the taxonomy~\cite{Arulkumaran2017DeepSurvey}, DRL reached state of the art performance in tasks such as 
robot control~\cite{Levine2016End-to-endPoliciesb} and board games~\cite{Silver2017MasteringKnowledge,Justesen2019DeepPlaying} where it even surpasses humans in some cases. Recent Deep Reinforcement learning (DRL) models, such as Deep Q-networks (DQN)~\cite{Mnih2013PlayingLearning,mnih2015humanlevel}, and Asynchronous Advantage Actor-Critic (A3C)~\cite{Mnih2016AsynchronousLearning}, learned to play video games with human level control using only images as input. As a result, they achieved human-level performances on Atari 2600 games~\cite{Bellemare2013TheAgents} such as breakout. Those models rely on the hypothesis that the optimal action can be decided based on a single frame.

However, these approaches operate on environments that can be totally observed (like a chess or GO board game), and not partially with a field of view which is smaller than the environment. To address this, an internal latent memory can be introduced~\cite{Hausknecht2015DeepMdps} to provide a space the model can use to store an approximation of the history of previous inputs and solve navigation problems~\cite{Mirowski2016LearningEnvironments,Zhu2017Target-drivenLearning,Oh2016ControlMinecraftb}, allowing learning in simulated environments such as Matterport3D~\cite{Chang2017Matterport3D:Environments},
\vizdoom~\cite{Kempka2016ViZDoom:Learningb,BeechingDeepSupercomputer}.

\subsection{Visual Analytics and Deep Learning}
 
Visual Analytics have been proven to be significantly helpful to deep learning (DL) experts to better understand their models~\cite{Hohman2019VisualFrontiers}, by providing insights on their decisions and inner representations. Such tools can be applied to Recurrent Neural Networks used as memory. In LSTMVis~\cite{Strobelt2017LSTMVis:Networks} users can formulate hypothesis on how the memory behaves with respect to the current input sentence. It displays memory elements in a parallel plot, and by selecting time intervals highlights the most active ones. The re-ordering of memory elements using a 1D t-SNE projection applied to handwriting trajectory prediction~\cite{Carter2016ExperimentsNetworkb} provides an overview of the representation and highlight patterns on how different feature dimensions reacts to different path \eg curvatures. 
Memory dimensions displayed over the input text of a character level prediction model~\cite{Karpathy2015VisualizingNetworks} highlights characters that trigger specific memory activations, and thus provide insights on how certain parts of the memory react to characters (\eg to quotes). RNN~evaluator~\cite{Ming2017UnderstandingNetworks}, uses clustering of memory elements into grids and associate them to word clusters for each input. This tool, also provides additional information on a chosen input in a detail on demand view. RetainVis~\cite{kwon2018retainvis}, a tool applied to the medical domain, studies how a modified model outputs its prediction based on data. With RetainVis, a user can probe an interesting data-point and alter it in a what-if approach to see how it affects predictions. To reach this level of interpretability, the model they used is altered, in a way that reduces its performances, which is different than our approach as visualize the model post-hoc. RNNbow~\cite{CashmanRNNbow:Networks}, is a tool able to handle different type of input domains, and can be adapted to DRL. However, RNNbow visualize the training of RNNs rather than their decisions. Such a tool, displays the gradients extracted from the model's training, and contextualize it with the input sequence and its corresponding output and label. In DRL with memory, the model does not receive a feedback at each decisions, but rather at the end of the game. This makes RNNbow more difficult implement as it produces large batches on which this tool have issues scaling to. As the authors mentioned, RNNbow targets non-experts user, and a domain specific tool may be required for experts.

As demonstrated by those tools, a decision at a time-step $t$ can be affected by an input seen at $t_{-n}$. In our case, such inputs are images and experts must first asses what the model did grasp from them before exploring what is stored in the memory. In addition the rewards from navigation tasks, are often sparse which results in a lack of supervision over actions, known as the credit assignment problem inherent to RL problems (the reward provided at a given time step can correspond to decisions taken at arbitrary time steps in the past). The model interacts with an environment it only sees partially, therefore, its performances can be altered by factors outside its inputs. This forces experts to visualize multiple time-steps in order to analyse a single decision which makes them more difficult to analyse with existing tools.

To our knowledge, DRL visualizations are under-represented in the literature compared to other methods on visualizing deep learning. LSTM activations from an A3C agent~\cite{Mirowski2016LearningEnvironments} have been displayed using a t-SNE~\cite{VanDerMaaten2008VisualizingT-SNE} projection. Despite being effective for an overview of the agent's memory, it offers limited information on the role of the memory. T-SNE projections have also been applied to memory-less DRL agents on 2D Atari 2600 games, in the seminal DQN paper~\cite{mnih2015humanlevel}, and in~\cite{Zahavy2016GrayingDqnsb}. DQNViz~\cite{Wang2018Dqnviz:Q-networks} displays the training of memory-less models under $4$ perspectives. First an overview of the complete training, action distribution of one epoch, a trajectory replay combined with metrics such as rewards and whether an action was random. DQNViz also includes a details view to explore CNN parameters. Such a tool, demonstrates the effectiveness of visual analytics solutions applied to DRL. However, DQNViz focuses on the training of the model, and how random decisions through training can affect it. In addition, the model of DQNViz is limited to fully observable 2D environments in which the only movements available are left or right and thus cannot be applied to navigation tasks. Finally, DQNViz is not designed to display or analyze any memory. 

In this paper, we address the under-explored challenge of visualizing a trained DRL model's memory in a 3D partially observed environment. We contextualize this memory with output decisions, inputs, and derived metrics. We also provide interaction to overview, filter, and select parts of such memory based on this context to provide clues on agents decision reasoning and potentially identify how the model uses its memory elements.











\section{Model and Design Goals}

This section presents the model we used to design and implement \tool. We describe the inner workings of those models and data characteristics. One key aspect being how the memory of DRL is created and updated by the agent, over space and time. Note that those data will be generated and then visualized with \tool\ after the training phase.

\subsection{DRL Model} 

The DRL model we relied on only receives pixels from an RGB image as input, from which it decides the action the agent should perform with the Advantage Actor-Critic (A2C)~\cite{Mnih2016AsynchronousLearning} algorithm. The model is composed of $3$ convolutional layers followed by a layer of Gated Recurrent Unit~(GRU)~\cite{Chung2014EmpiricalModeling}, and Fully Connected (FC) layers to match the actions set $A$. This model is inspired by \emph{LSTM A3C} as presented in~\cite{Mirowski2016LearningEnvironments} with A3C instead of A2C, and a LSTM~\cite{Greff2017LSTM:Odyssey} instead of GRU. Those changes reduce the agent's training time, while preserving its performances. The underlying structure that allows our model to associate raw pixels to an action is illustrated on Fig.~\ref{fig:overview} and described as follows:

\begin{description}[leftmargin=0pt]

    \item [Stage 1: Environment $\rightarrow$ Image.]
    First, the agent's field of view is captured as image $x_t$, i.e. a matrix with dimensions of $112 \times 64$ with $3$ RGB color channels.
    
    \item [Stage 2: Image $\rightarrow$ Feature vector.]
    The image $x_t$ is then analyzed by $3$ convolutional layers designed to extract features $f_t$, resulting in a tensor of $32$ features shaped as a $10 \times\ 4$ matrices. These features are then flattened and further processed with a Fully Connected (FC) layer. Formally, the full stack of convolutional and FC layers is denoted as function $f_t = \Phi(x_t,\theta_{\Phi})$ with trainable parameters $\theta_{\Phi}$ taking $x_t$ as input and given features $f_t$ as output.
    
    \item[Stage 3: (Features + previous memory) $\rightarrow$ New memory.]
    The model maintains and updates a latent representation of its inputs using a Gated Recurrent Unit (GRU)~\cite{Chung2014EmpiricalModeling}, a variant of recurrent neural networks. This representation, called hidden state $h_t$, is a time varying vector of $128$ dimensions, which is updated at each time-step $t$ with a trainable function $\Psi$ taking as input the current observation, encoded in features $f_t$, and the previous hidden state $h_{t-1}$, as follows: $h_t = \Psi(h_{t-1},f_t,\theta_{\Psi})$. 
    
    \item[Stage 4: Memory vector $\rightarrow$ Action.] 
    The model maps the current hidden state $h_t$ to a probability distribution over actions $A$ using a fully connected layer followed by a softmax activation function, denoted as the following trainable function:
   
    $a_t = \xi(h_t,\theta_{\xi})$ with trainable parameters $\theta_{\xi}$.
    The highest probability corresponds to the action $a_t$ which the agent estimated as optimal for the current step $t$.
\end{description}

The full set of parameters $\theta = \{\theta_{\Phi}, \theta_{\Psi}, \theta_{\xi}\}$ is trained end-to-end. We used $16$ parallel agents and updated the model every $128$ steps in the environments. The gamma factor was $0.99$, and we used the RMSProp~\cite{tieleman2012lecture} optimizer with a learning rate of $7e-4$. We trained the agent over $5$M frames, with a frame skip of $4$. During training, the agent does not necessary pick the action with the highest probability, as it needs to explore its environment, and eventually find better solutions. However, once the agent is trained, it always chooses the action with the highest probability. 

\begin{figure}[t!]
    \centering
    \includegraphics[width=8cm]{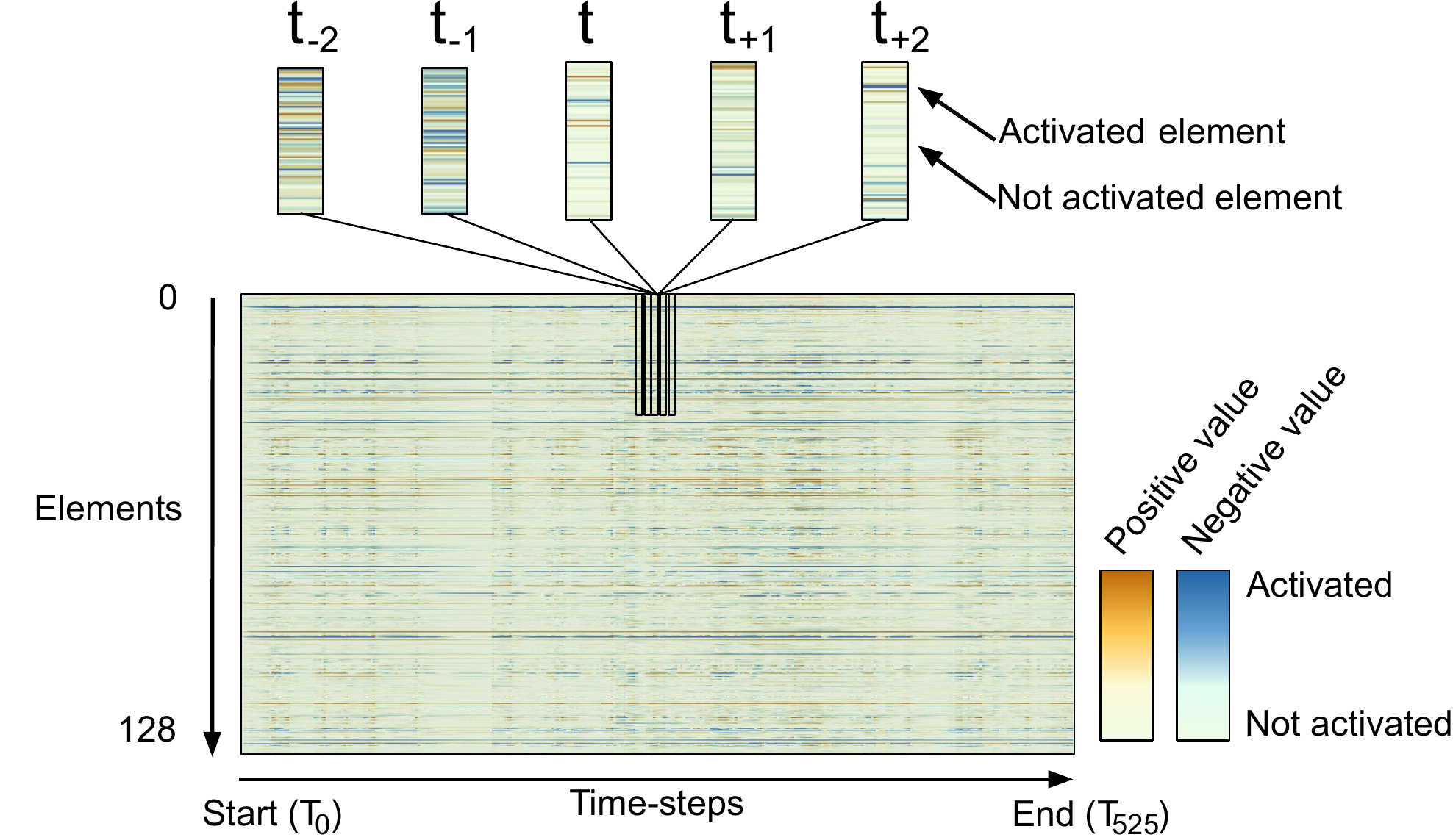}
    \caption{Memory construction process: at a current time-step $t$, the agent updates its memory by producing a new memory vector. Each dimension of this vector (represented as a column) is appended to the previous ones chronologically (from left to right). As a result, each row of the appended vectors represent the actions of a single memory element.}
    \label{fig:matrix}
    \vspace{-.5cm}
\end{figure}

\subsection{Constructing the Memory of DRL}

In the partially observed navigation problem we focus on, the agent only sees the current observation, \ie what is in its field of view at the time-step $t$. However, past observations are also relevant for decision making (\eg to gather previously seen items). Therefore the agent needs to build a representation of relevant information extracted from the history of observations. This information is encoded in $h_t$, a high dimensional ($128$ in our case) time varying vector. 

Fig.~\ref{fig:matrix} represents the construction process of the hidden states matrix, which consists of the hidden states $h_t$ over the time of an episode --- the central visualization in \tool\ (Fig.~\ref{fig:teaser}). Each hidden state is vertically aligned per time-step $t$ at which they are produced. Therefore, the accumulation of hidden states forms a large 2D matrix, where the horizontal axis is time ($h_{t-1} < h_{t} < h_{t+1}$) and the rows are elements. A row of this 2D matrix represents the evolution and activity of a hidden state element through time and space as the agent moves. The activity of a hidden state element is characterized by its value. In our case, each element of the hidden states is a quantity within the range $[-1,1]$. A value close to $0$ represents low activity, whereas a value close to any extremity represents high activity. As it can be seen in Fig.~\ref{fig:matrix}, hidden states can drastically change their values between two time-steps. Such value changes can be widely observed across hidden states elements during episodes. However, it remains unclear which elements, correspond to which representations, and thus, responsible for decisions. 

Norms of latent activations are an informative way of visualizing influences~\cite{cadene2019murel,zhou2016learning}. With modern training methods such as weight decay, dropout and batch normalization, it is highly improbable that a high activation can occur for unused features. An alternative to hidden state activations would be to analyze gradients of action probabilities with respect to hidden states~\cite{selvaraju2017grad,CashmanRNNbow:Networks}. Such an approach can provide information on how a chosen action is directly tied to the current state of the memory, and which dimension influences the more this decision. However, in \tool\ we focus on actions through the episode as a sequence rather than small sub-sequences. When visualizing activations of an LSTM on text, Karpathy et al.~\cite{Karpathy2015VisualizingNetworks} discovered a pattern occurring outside back-propagation limitations of the gradient signal. A solution would be to display both activations and gradients, however preserving the usability and interpratbility of a tool conveying such information is challenge yet to be tackled.

\section{Design of \tool}

We built \tool\ as a visual analytics interface to understand the connections between the latent memory representation (as depicted in Fig.~\ref{fig:matrix}) and decisions of an agent trained using Deep Reinforcement Learning. \tool\ primarily exposes the internal memory (Fig.~\ref{fig:teaser}) which is interactive and provides \emph{overviewing}, \emph{filtering} and \emph{reduction} both for exploration and knowledge generation~\cite{Keim2008VisualChallenges}. \tool\ is designed towards experts in DRL to identify elements responsible for both low-level decisions (\eg move towards a spotted HP) and eventually higher-level strategies (\eg optimizing a path).

\subsection{Design Motivation and Goals}
\label{sec:challenge}

We iteratively built \tool\ with frequent meetings from colleagues experts in DL and DRL (a total of $12$ meetings with three experts over $7$ months). We first identified their current process to analyze trained agents, \eg recording videos to playback agents episodes (from its point of view) and decisions (actions probability) to get a sense of the strategy. We also observed experts do a system print of the models' inner values, sometimes add conditions to those prints (\eg when an item is in the field of view of the agent), and manually look for unusual values. Our approach was to re-build a similar interface in \tool\ with input/output views and facilitate playback, but 1) in an interactive way, and 2) by adding advanced, coordinated views to support advanced models visualization aimed at models developers~\cite{Hohman2019VisualFrontiers} (\eg view on the agent's memory).

Based on a review of the current practices of researchers from our focus group, and related work, we identified the following design goals (\textbf{G}) to be addressed to understand the behavior of a trained agent using a learning model for navigation problems:

\begin{enumerate}[noitemsep,label=\textbf{G\arabic*},align=left]
    
    \item \label{R:context} \textbf{Show an agent's decisions over (a) space and (b) time}, especially input and outputs of the model.

    \item \label{R:memory} \textbf{Expose the memory's internal structure}, \ie the temporal vector built over time (Fig.~\ref{fig:matrix}).
    
    \item \label{R:compare} \textbf{Link memory over (a) time and (b) decisions} with multiple endpoints, \eg starting from any time, location, memory or trajectory point.

    \item \label{R:annotate} \textbf{Filter a sub-set of the memory (a sub-space)} tied to a specific agent behavior or strategy.
    
\end{enumerate}


\subsection{Overview and Workflow of \tool}

Fig.~\ref{fig:teaser} shows an overview of \tool\ where the most prominent visualization is the memory timeline of a trained agent (\ref{R:memory}). The primary interaction is browsing the timeline and playback the input video feed and action probabilities (\ref{R:context}). Beyond re-playing scenarios, \tool\ implements multiple interactions to:

\begin{enumerate}[noitemsep]

    \item \emph{Overview} the memory and check what the agent sees and its decisions; visual cues for selection are dark, consecutive patterns (Fig.~\ref{fig:matrix}).
    
    \item \emph{Filter} the timeline when something is of interest, \eg related to the activation, but also with additional timelines (actions, etc.).
    
    \item \emph{Select} elements whose activation behavior is linked to decisions. Those elements are only a subset of the whole memory and are visible on Fig.~\ref{fig:teaser}~\ding{195}.

\end{enumerate}

Those interactions are carried out using a \emph{vertical thumb} similar to a slider to explore time-steps $t$ and select intervals. Such a selection is propagated to all the views on the interface, whose main ones are \emph{image (perception)} and \emph{probabilities (of actions)} which provide context on the agent's decisions~(\ref{R:context}~(b)). The input image can be animated as a video feed with the playback controls, and a saliency map overlay can be activated~\cite{Springenberg2014StrivingNetc,greydanus2017visualizing} representing the segmentation of the image by the agent. The \emph{trajectories} view (Fig.~\ref{fig:teaser}) displays the sequence of agent positions $p_{t-1} > p_{t} > p_{t+1}$ on a 2D map~(\ref{R:context}~(a)). This view also displays the items in the agent's field of view as colored circles matching the ones on the timeline. The position $p_t$, and orientation of the agent are represented as an animated triangle. The user can brush the 2D map to select time-steps, which filters the memory view with corresponding time-steps for further analysis~(\ref{R:compare}~(a)). \tool, also includes a \emph{t-SNE}~\cite{VanDerMaaten2008VisualizingT-SNE} view of time-steps $t$ using a two-dimensional projection (Fig.~\ref{fig:teaser} bottom left). T-SNE is a dimensionality reduction technique, which shows similar items nearby, and in this view, each dot represents a hidden state $h$ occurring in a time-step $t$. The dot corresponding to the current time-step $t$ is filled in red, while the others are blue. The user can select using a lasso interaction clusters of hidden states to filter the memory with the corresponding time steps. Dimensions among the selected hidden states can then be re-ordered with any criteria listed in Table~\ref{table:sort}, and brushed vertically (Fig.~\ref{fig:teaser}~\ding{175}).

The result of such an exploratory process is the subset of elements of the memory (rows) that are linked to an agent's decision (Fig.~\ref{fig:teaser}~\ding{175}). This addresses the design goal~\ref{R:annotate}. Such subset can be seen as a memory reduction which can be used as a substitute to the whole memory (we will discuss it in the perspective sections). This subset can also be used in other episodes listed as clickable squares at the bottom left corner of \tool.

\subsection{Memory Timeline View}

The \emph{memory timeline} exposes the memory's internal structure~(\ref{R:memory}), which is vector (vertical column) of $128$ dimensions over $525$ time-steps as a heat-map (Fig.~\ref{fig:teaser}~\ding{173}) from which an interval can be brushed for further analysis. Each cell (square) encodes a quantitative value, whose construction is illustrated in Fig.~\ref{fig:matrix}, using a bi-variate color scale from~\cite{Light2004TheGraphics} with blue for negative values and orange for positive values. Preserving the values as they were initially produced by the model is pertinent as some memory elements (rows) can have both positive and negative values, which may not have the same signification for the model and thus cause different decisions. This will be further explored in Sec.~\ref{sec:thom}.

By default \tool\ displays the vector as it is produced by the model, hence the order of elements has no particular semantic. The memory can be re-ordered using a drop-down menu according to comparison criteria listed in table~\ref{table:sort}. With the \textsc{activation} criteria, a user can observe elements that may be most involved in decisions, while with \textsc{change}, elements that may be the most used by the model are emphasized, with \textsc{similar}, a user can see elements with constant activations during selected intervals.
In addition of those criteria, we provided the possibility to re-order the memory as presented in~\cite{Carter2016ExperimentsNetworkb} \ie a one dimensional t-SNE projection of the absolute values. The re-ordering can either be applied to the whole memory or a selected interval. An order is preserved across memory filters and episodes until the user changes it.

\begin{table}[t!]

\centering
\begin{tabular}{ p{1.2cm} p{2.9cm}  p{3cm} }
Criteria & Formula & Description \\  
\hline

\textsc{activation}  & $ \sum\limits_{t=1}^{n} |h_{ti}|$ & Elements most involved in decisions. \\  

\textsc{change} & $ \sum\limits_{t=1}^{n-1} |h_{ti} - h_{t+1i}| $ & Maximal change. \\ 

\textsc{stable} &  \textsc{change}$^{-1}$ & Minimal change. \\

\textsc{similar} & $|\frac{1}{n} \sum\limits_{t=1}^{n-1} h_{ti} - \frac{1}{k}  \sum\limits_{t=1}^{k-1} h_{tj}|$  & Elements in average different during an interval of $k$ time-steps than outside it.\\

\end{tabular}
\caption{List of re-ordering criteria as they appear in \tool. $t$ is the current time-step, $n$ the number of steps ($525$ at most), and $i$ the element.}
\vspace{-.2cm}

\label{table:sort}
\end{table}

\subsection{Derived Metrics View}

\begin{table}[t!]

\centering
\begin{tabular}{l c l}
Metric & Data Type & Values  \\  
\hline
\emph{Health of the agent} & Quantitative & death \textbf{[0,100]} full  \\ 
\emph{Event (item gathered)} & Flag & \textbf{(1)}  gathered \\ 
\emph{Item in FoV} & Binary & no item \textbf{(0, 1)} item  \\ 
\emph{Orientation to items} & Degree & left \textbf{[-45,45]} right \\  
\emph{Variation of orientation} & Quantitative & stable \textbf{[0,30]} change \\
\emph{Decision ambiguity} & Ratio & sure \textbf{[0,1]} unsure
\end{tabular}
\caption{List of \emph{derived metrics} (from top to bottom on Fig.~\ref{fig:teaser}~\ding{174})}
\label{table:metrics}
\vspace{-.5cm}
\end{table}

The derived metrics timeline addresses the design goals~\ref{R:compare} and~\ref{R:annotate}. It represents metrics calculated from ground truth information provided by the simulator. Those metrics aim at supporting the user in finding interesting agent behaviors such as \emph{What does a trained agent do when it has no health pack in its field of view?}. The view encodes measures of both the inputs (\eg health pack is spotted) simulator (\eg reward) and outputs (\eg actions). Finally \tool\ features a stacked area chart of actions probabilities encoding probabilities per action represented by colors corresponding to the ones on the action distribution graph in~\ref{fig:teaser}. With this visualization, users can observe similar sequences of decisions.

\begin{figure}[t!]
    \centering
    \includegraphics[width=8cm]{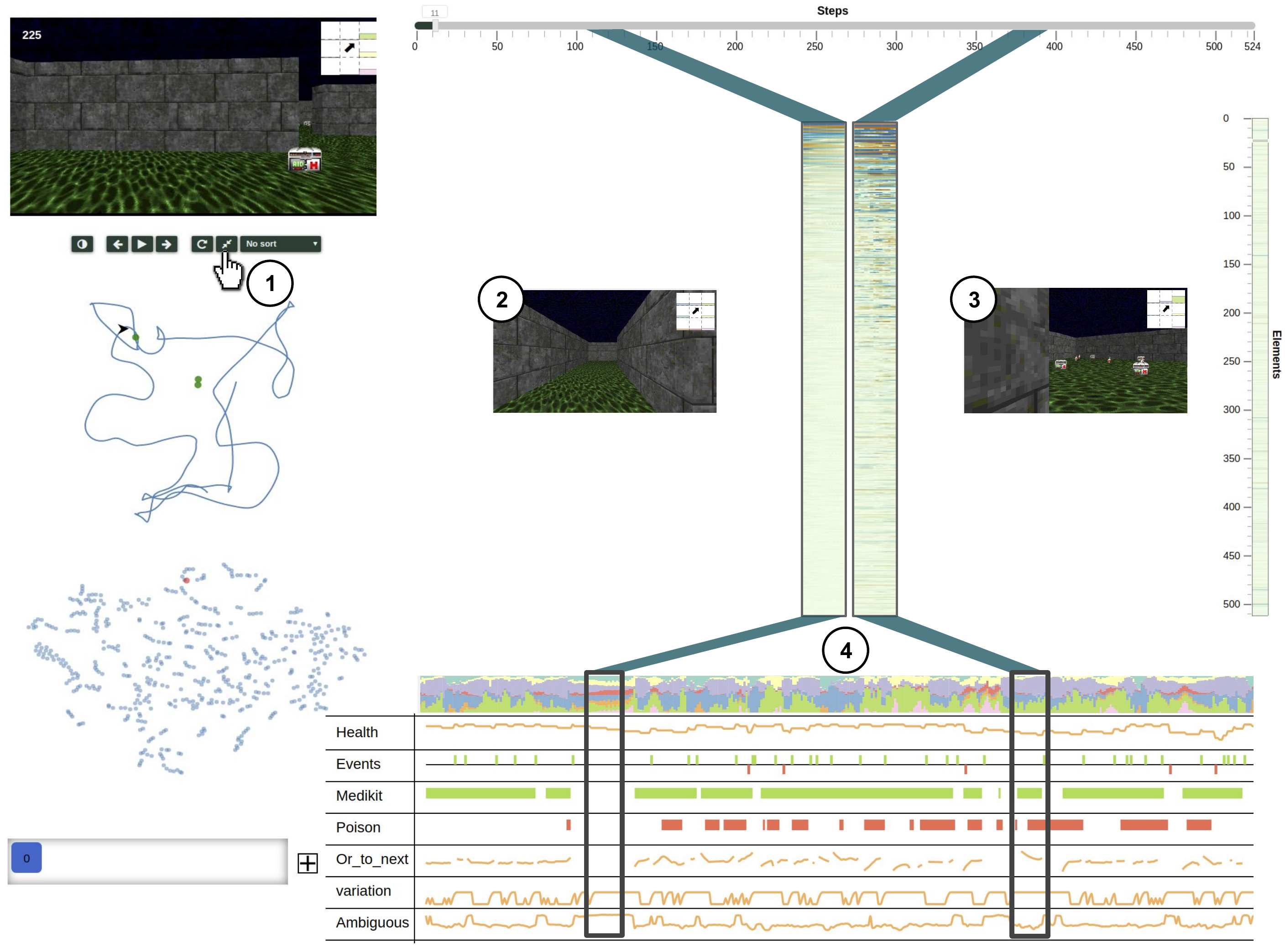}
 
    \caption{ \tool\ allows to compare selected time intervals~\ding{172}. For instance to compare when agents face dead-ends~\ding{173} and when they face health-packs~\ding{174}. One can observe that more elements are active while the agent is facing HPs than while facing a dead-end. Perhaps those elements are encoding information concerning HPs. When facing a dead-end, both the orientation variation and decision ambiguity are high which can be interpreted as the agent hesitating on which action to choose.
    }
    \label{fig:metric}
    \vspace{-.5cm}
\end{figure}

The derived metrics and stacked area chart are below the memory timeline and share the vertical thumb from the memory slider~(\ref{R:compare}~(a)) to facilitate comparisons between the memory and the behavior of the agent (\ref{R:compare}~(b)) as depicted in Fig.~\ref{fig:metric}. The derived metrics can be dragged vertically by the user as an overlay of the memory to compare metrics with activation values, and identify memory elements related to them~(\ref{R:annotate}). A constant activation of an element during the same intervals of a metric such as \emph{HP in FoV}, while being different otherwise; may hint that they are related. We provide a full list of metrics in table~\ref{table:metrics}. Two metrics are particularly complex and described as follows:

\begin{description}[leftmargin=0pt]
    \item[\emph{Variation}] describes how the the agent's orientation (\ie its FoV) changes over three consecutive time-steps. High variations indicate hesitation in directions and intervals during which the agent turns around, whereas low variations indicate an agent aligned with where it wants to go. However, in some cases (\eg the agent stuck into a wall), actions may have no effect on the agent's orientation which lead the variation to remain low.
    
    \item[\emph{Ambiguity}] of a decision is computed using the variance $V$ of action probabilities. The variance describes how uniform actions probabilities are with respect to the mean. A variance $V = 0$ indicates that there is no difference between actions probabilities, and hence that the agent is uncertain of its decision. In the other way, a high variance represents strong differences in the actions probabilities and the agent's high confidence on its decision. Since the sum of all actions probabilities equals to $1$, the variance is bounded within the range $[0,1]$. To ease the readability, the variance is inverted as it follows: $ambiguity = 1-V$. An ambiguity close to $1$ represents an incertitude in the agent's decision.
  
\end{description}

\section{Implementation}
\label{sec:implementation} 

To explore the memory of a trained agent, one needs to create \emph{instances} of exploration scenarios. For experts evaluations (Sec.~\ref{sec:eval}) we used a trained agent to explore $20$ times the environment with different configuration (\ie positions of items, start position of the agent, its orientation). During those episodes, we collected at each time-step information from the agent such as its FoV image, action probabilities, memory vector, and information from the environment such as the items in the agent's FoV, the position of the agent, the agent's orientation and its health. The collected data is formatted as a JSON file which groups data elements per episodes and then per steps with an average of 30Mo per episode. Those data are generated using DRL models implemented in Pytorch~\cite{Paszke2017AutomaticPyTorch}, and formatted in Python 3. More technical details are provided as supplemental material.

The user interface of \tool\ loads data using JavaScript and D3~\cite{Bostock2011D3:Documents}. The interactions between the model and the front-end are handled by a Flask Python server. The data, separated per episode is generated in a plug-in fashion \ie without altering the model nor the simulator. Both the interface code source~(\url{https://github.com/sical/drlviz)} and an limited interactive prototype~(\url{https://sical.github.io/drlviz)} are available online.

\begin{figure*}[ht!]
    \includegraphics[width=17cm]{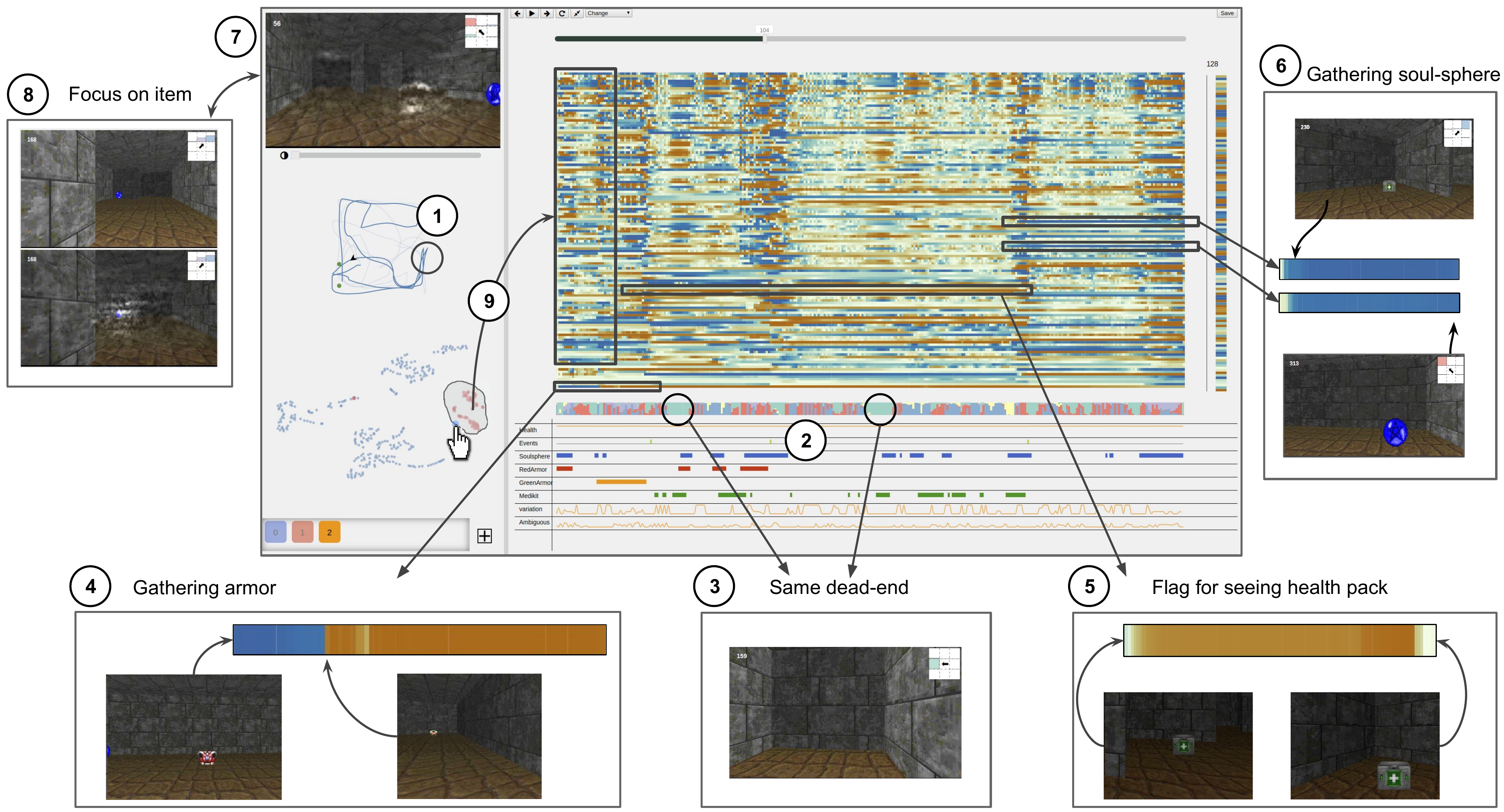}
    \centering
    \caption{ 
    Summary of the insights gained by the experts. \edward\ noticed two intervals during which the agent only turned right, by using both trajectory~\ding{172} and stacked area chart of actions~\ding{173} views. Once he replayed those sequences, he stated that the agent came twice in the same dead-end~\ding{174}. \Thomas\ observed a hidden state dimension which is blue when the agent sees the red armor before the green armor, and then remained orange until when he saw the green armor~\ding{175}. \Arthur\ probed a dimension that is active as the agent first saw the HP, and remained active until it gathered it. \Edward\ also identified two hidden state elements that changes as the agent gathered the health pack and then kept their values until the end of the episode~\ding{177}. Using saliency maps~\ding{178}, \arthur\ observed that the agent ignored the soul-sphere until it gathered the first three items~\ding{179}. Finally, \thomas\ identified clusters in the t-SNE projection which corresponds to the agent's objectives \eg gathering the green armor~\ding{180}. 
    } 
    \label{fig:eval1}
    \vspace{-.5cm}
\end{figure*}

\section{Experts Evaluation}
\label{sec:eval}

We conducted a user study with three DRL experts who are experienced researchers building DRL models and match the target profile for \tool~\cite{Hohman2019VisualFrontiers}. We report on their use of \tool, as well as insights they gathered. Those results may not be confirmed or denied using \tool, but provide hints to formulate hypothesis that can then be studied outside \tool\ \eg through statistical evidence.

\subsection{Protocol and  Navigation Problem}

We recruited three DRL experts (\edward, \arthur, \thomas) from two different academic laboratories to evaluate \tool.
They were shown a $10$ minutes demonstration of \tool\ on a simple \vizdoom\ scenario: \emph{health gathering supreme}. The evaluation started with \tool\ loaded with data extracted from a model developed by \edward, and ended after $35$ minutes, during which experts could explore the displayed data. While using \tool\ experts were told to explain their thoughts and what they wanted to see. Then, experts were asked to fill a post-study questionnaire to collect their first impressions with open questions such as \emph{"Which part of \tool\ was the least useful?"}. The evaluation ended with a discussion guided by the post-study questionnaire on their experience using \tool\ and how it can be improved. The complete evaluation lasted in average $1$ hour depending on the length of the discussion. The model used was an A2C~\cite{Mnih2016AsynchronousLearning} with 3 convolutional layers and GRU layer with $128$ dimensions as memory.


\subsection{Feedback from \Edward}

\edward\ is the most experienced expert for this evaluation as he designed both the model and the navigation task~\cite{BeechingDeepSupercomputer} and created animations of agents behaviors. \Edward\ was our primary collaborator to design and build \tool.

Fig.~\ref{fig:eval1} shows \tool\ loaded with the \emph{k-item} scenario. \Edward\ first selected an interval corresponding to the agent searching and gathering the last item. This interval started one step after the agent gathered the HP (third item), and ended as the agent gathered the soul-sphere (last item). \Edward, then used the \textsc{change} criteria to re-order the interval. While replaying it, he noticed two elements with similar activations~\figref{fig:eval1}{177}. Those elements remained blue during the interval, however they were inactivated (gray) during the rest of the episode. With further investigation, \edward\ noticed that those elements were active $4$ steps before the agent gathered the HP. \Edward\ described those elements as \emph{flags} \ie elements that encodes binary information. \Edward's intuition was that the agent learned to complete the navigation problem by focusing on one item at the time. And only used its memory to encode information on items it already gathered, and hence which item it should currently gather. \textbf{\Edward\ concluded that the two elements may be the agent's representation that it gathered the HP, and hence that it should now focus on gathering the soul-sphere.}

Then using the action probabilities temporal stacked area chart~\figref{fig:eval1}{173}, \edward\ noticed a specific time interval during which the agent repeated the same action for almost $15$ steps. Intrigued by such behavior, \edward\ replayed this interval and noticed that the agent was within a dead-end~\figref{fig:eval1}{174} and repeated the action \emph{right} until it changed its orientation up to $180$ degrees. \Edward\ commented that observing such interval is interesting because as the agent converges towards an optimal policy, it may have less chances to encounter dead-ends, and thus forgot how to escape them. \Edward\ also observed a similar interval with only \emph{right} actions in which the agent escaped the same dead-end. \textbf{\Edward\ concluded that the dead-end was not encoded in the agent's memory, and hence the agent returned to it while searching for items.} 
 
\subsection{Feedback from \Arthur}

Our second expert, \arthur, started by re-ordering the memory using the \textsc{stable} criteria. He noticed a hidden state element, and zoomed (vertical brush) on it. This element had continuous activations starting as the agent first saw the HP and remained active until the agent gathered both the red armor and the HP. \textbf{Because such element is active regardless of the items the agent gathered yet, \arthur\ interpreted this element as a flag encoding if the agent has seen the health pack or not.}

Then \arthur\ focused on the saliency maps combined with episode playback. He noticed that in one episode, the agent encountered the soul-sphere (last item) before it gathered the red armor (second item). During those time-steps, the saliency maps are not activated towards the soul-sphere despite being the agent's FoV~\figref{fig:eval1}{178}, and the memory had no visible changes. \Arthur\ intuition was that the agent did not perceived the item. In the final steps of the episode, once the agent gathered the firsts $3$ items and then re-encountered the soul-sphere, the saliency maps were activated towards it~\figref{fig:eval1}{179} and the memory activations changed. \Arthur\ expressed that \emph{"It is interesting because as soon as it sees it [the soul-sphere] its behavior changes"}. \textbf{\Arthur\ concluded that the agent intentionally ignored the soul-sphere before it gathered previous items, and as \edward\ mentioned, the agent learned to solve this navigation problem by focusing on one item at a time.}

\subsection{Feedback from \Thomas} 
\label{sec:thom}
\thomas\ began his exploration with the t-SNE 2D projection as entry point to identify clusters of hidden states. \Thomas\ selected groups of states using the lasso selection \figref{fig:eval1}{180} to filter the memory timeline. The selected cluster represented consecutive steps, forming a continuous time interval. After replaying this interval, \thomas\ observed that it started at the beginning of the episode and ended when the green armor (first item) entered the agent's FoV. \textbf{\Thomas\ interpreted this cluster as corresponding to an agent objective, in this case gathering the first item.}

Following up on the previously identified cluster, \thomas\ re-ordered it with the \textsc{stable} criteria. \Thomas\ noticed one particular hidden state dimension that was activated in blue until the green armor entered the agent's FoV, and then was activated in orange for the rest of the episode. \Thomas\ interpreted such element activation as a flag encoding if the agent has seen the green armor. However, after observing this element activations across episodes, \thomas\ noted that it was inactivated (grayish) at the start of an episode. After re-playing this episode he observed that the agent had no armor in its FoV, as opposed to the first episode analyzed where the agent started with the red armor in its FoV. In another episode, where the agent has the green armor in its FoV since the start, the element was constantly activated in orange. \textbf{\Thomas\ concluded that this element encoded if the agent saw an armor rather than just the green armor.} However, once the agent gathered the green armor, the element remained orange despite still having the red armor in the agent's FoV. \textbf{\Thomas\ added that this element also encodes if the agent gathered the green armor.}

\section{Discussion}

In this section, we discuss the collected feedback from experts, as well as the limits of the current version of \tool. 

\subsection{Summary of Experts Feedback}

Experts filled a post-study questionnaire relative to \tool\ usefulness and usability. Overall \tool\ was positively received by all them: both \edward\ and \arthur\ stated that \tool\ is "\emph{interesting to explain the behavior of the agent}" and "\emph{easy to use}". However, \thomas\ stated that he felt "\emph{overwhelmed at first, but soon got used to navigation}". All 3 experts evaluated the 2D t-SNE projection as the most useful view because it can provide insights on the agent's memory and strategies. They used this view as entry point on at least one episode. They commented that the re-ordering was effective to observe desired hidden states dimensions. Both \arthur\ and \thomas\ used the \textsc{stable} criteria because it highlights elements that are different from the rest and should correspond the selected interval. In the other hand, \edward\ preferred the \textsc{change} re-ordering criteria because those elements have information concerning the interval. \Thomas\ also noted that "\emph{it´s handy being able to drag it up [derived metrics timeline] and overlay it on the hidden states}" (\ref{R:compare}). The experts concluded that the agent learned to solve this task sequentially, \ie by focusing on gathering one item at the time. And thus that the agent only stored information corresponding to which items its has gathered rather than the positions of every seen items at any time-steps.

All three experts evaluated the memory reduction interaction that filters the memory view (zoom) not intuitive and hard to use without loosing visual contact with the hidden state dimensions they wanted to focus on. This partially validates our memory reduction goal (\ref{R:annotate}). On this matter, \edward\ commented that since this agent's memory has $128$ dimensions the zoom is not as useful as it could on larger memories. \Arthur\ also commented on the use of the different re-ordering criteria, and that their specific functioning was hard to understand, especially the projection. \Thomas\ also mentioned that he "\emph{doesn't fully understand how the projections re-ordering methods are helpful}". To tackle those issues, \thomas\ suggested to use the derived timeline to re-order the memory, \ie observe hidden states activations when a feature enters the FoV. \Thomas\ also commented that a horizontal zoom could be useful to focus on one particular time interval, and reduce the number of steps to observe. \Edward\ mentioned that brushing the memory while keeping activation areas as \emph{squares}, \ie both horizontally and vertically could be a better way to implement a more consistent zooming interaction.

\subsection{Limits} 
\emph{Generalization} and \emph{scalability} are the main limits of the current version of \tool. Regarding generalization, specific calculations need to be made such as for the derived metrics timeline that is generated from the simulator \ie the items in the agent's FoV. So the current metrics are tied to \vizdoom\ but minor adaptation of the tool to specific environments will be needed, but requiring technical knowledge. In the next section we will explain how the interaction techniques in \tool\ can be used beyond the tool for better timeline comparisons. Scalability is always a concern with visualization techniques. \tool\ supports long episodes and variable memory size. However, if those are larger than the screen real estate (\eg beyond on average $1200$ steps and more than $1000$ memory dimensions) each memory cell would be smaller than one pixels, and thus difficult to investigate. To tackle such an issue, LSTMVis~\cite{Strobelt2017LSTMVis:Networks} introduced a parallel coordinate plot with each line encoding a memory element. However, with \tool\, we sought to support trend detection and thus encode the memory overview using colored stripes~\cite{fuchs2013evaluation} which complies with our data density challenge and requirement to align the memory with the derived metrics below. We then rely on zoom interactions for details for both time and elements.

We plan in the future to support aggregation strategies~\cite{Wongsuphasawat2011LifeFlow:Sequences} to provide more compact representation of the timelines. Alignment by event of interest~\cite{DavidWang2008AligningRecords} (\eg gathering an item) may also provide more compact representations of metrics, and also better support comparison of behavior before and after this event. A concern raised by experts was the communication of insights gained during the exploration process. We plan to explore summarizing techniques for those insights, such as state charts~\cite{Salomon2018HumanAutomata} in which each state corresponds to a local strategy \eg reach an item. 

\section{Perspectives}
\begin{figure*}[ht!]
    \includegraphics[width=14cm]{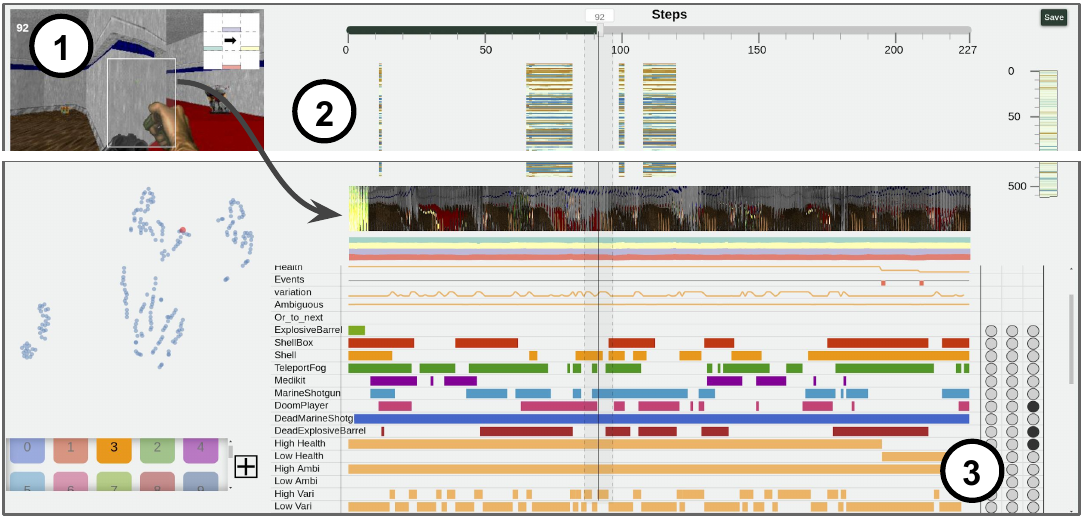}
    \centering
    \caption{Extended version of \tool\ loaded on with death-match data. From a slit square selection \ding{172} outputs a timeline that summarizes the agent's point of view~\ding{173}. And the additional metrics and operators~\ding{174}. } 
    \label{fig:timelineX}
    \vspace{-.5cm}
\end{figure*}

We present and discuss three works in progress that may be potential improvements of \tool, based on experts feedback, that primarily expand its exploration power and generalization.

\subsection{Memory Reduction} 
\label{sec:reduc}

\begin{table}[b!]

\centering
\begin{tabular}{ l l l l l }
Type of reduction  & Steps   & HP  & Poison & Health \\  
      & survived & gathered &  gathered &  \\  
\hline

Full-memory  & \textbf{503.98} & 37.56 & \textbf{4.28} & 81.47 \\  

Half-memory & 493.92 & \textbf{37.88} & 4.66 & \textbf{81.61} \\ 

\end{tabular}
\caption{Performances of agents with different memory reduction strategies (each averaged over $100$ episodes).
} 

\label{table:perf}

\end{table}
As experts noticed during interviews, agents memory can often be sparse (\eg Fig.\ref{fig:teaser}) or hold redundancy (\eg Fig.~\ref{fig:eval1}~\ding{177}). Thus we hypothesize that some elements may either never be activated or there might be multiple, redundant activation at the same time. We conducted an experiment to assess that some sub-set of the memory is sufficient to solve a navigation problem, and the rest may be discarded. We used the \emph{health gathering supreme} scenario in which the agent must collect HPs to survive, hence it is easier to solve than k-item. With a larger memory of $512$ dimensions, we "removed" hidden state elements by multiplying them by $0$ during the experiment.

Table~\ref{table:perf} shows similar performances between agents with full and top half memory re-ordered with \textsc{activation} on the \emph{health gathering supreme} scenario and a large $512$ memory of $512$ dimensions. One hypothesis to draw is that the agent has at least $256$ non-essential elements. Efficient selection of those elements remains a challenge, as it must account complex temporal dependencies. We built an explorable visualization~\cite{Jaunet:2019} to support this process manually and implemented several strategies, to compare with agents behavior (\eg being hesitant and producing mistakes such as running in circles or bumping into walls which could have been avoided using its full memory). This paves the way for direct filtering by elements of the memory heat-map in a future version of \tool, as the current version only selects temporal intervals. 







\subsection{Guiding Exploration with Extended Timelines}

During our interviews, experts suggested to better support the memory analysis process, as the current version of \tool\ relies on visual exploration by the user, with no specific guidance. We identified two areas of improvement for a future version of \tool: adding more metrics, and advanced filtering. Regarding the metrics, table~\ref{table:metrics} introduced derived indicators from the agent decision. Fig.\ref{fig:timelineX} illustrate that more metrics can be added using variations of their parameters (\eg changing variability thresholds or the distance to consider an enemy is in the FoV or not) which support more questions a user may want to investigate. Such metrics are represented in a compact way, and easy to activate by scrolling down, while remaining focused on the memory. Regarding the comparison, the current version only implements \emph{juxtaposition} and \emph{overlay}; while \emph{explicit encoding}~\cite{gleicher_visual_2011} is a third way to compare timelines and memory. We applied this third way by adding a boolean queries builder~\cite{lex2014upset} using \texttt{AND} or \texttt{OR} to filter timelines. Those boolean operators are also applicable to all views of \tool, such as 2D-map, t-sne or a brush on the memory. This helps users to combine multiple views and answer question such as \emph{Where are the areas of memory with the agent has high health, in this part of the environment, with an enemy and an explosive barrel in FoV?}. This results into intervals in which the agent is susceptible to shoot on barrels to kill enemies.

In order to summarize the input images and ease their comparison with derived metrics, we developed \emph{slit square} interaction based on slit tears~\cite{tang2009exploring}. With slit square, a user can brush a square on the inputs. Those squares are then compacted to with the width shared by all time-aligned elements in \tool.

\subsection{Generalization to other Scenarios and Simulations}
\label{sec:generalization}


Finally, we started investigating using \tool\ as a general-purpose tool for any trained agents with a memory and spatio-temporal informations. Fig.~\ref{fig:timelineX} illustrates \tool\ loaded with a different scenario where the agent shoots towards enemies on the  death-match~\cite{Lample2017PlayingLearning} with the Arnold model. In general, \tool\ can be used beyond \vizdoom\ (\eg referred in~\cite{BeechingDeepSupercomputer}), such as Atari games~\cite{Wang2018Dqnviz:Q-networks} without any major change. Using pixel-based representations~\cite{keim_designing_2000} and zooming~\cite{kerpedjiev2018higlass} would assure scalability of the timeline representations with scenarios requiring more time steps. We plan to conduct further research to identify other metrics and extend \tool\ to other simulators mentioned by our experts, such as Matterport3D~\cite{Chang2017Matterport3D:Environments} and Habitat-AI~\cite{Savva2019Habitat:Research} for real-world scenarios, and competitions such as Animal-AI~\cite{crosby2019animal}. 


\section{Conclusion}

In this work, we introduced \tool, a visual analytics interface which allows users to \emph{overview}, \emph{filter} and \emph{select} the memory of Deep Reinforcement Learning (DRL). Analysts using \tool\ were able to explain parts of the memory of agents trained to solve navigation problems of the \vizdoom\ game simulator, in particular local decisions and higher level strategies. \tool\ received positive feedback from experts familiar with DRL models, who managed to browse an agent memory and form hypothesis on it. \tool\ paves the way for tools to better support memory reductions of such models that tend to be large and mostly inactive. 

\section{Acknowledgement}
This research was partially funded by the M2I project \url{http://www.mob2i.fr/}, Projet Investissement d'Avenir on urban mobility by the French Environment Agency (ADEME).
\bibliographystyle{eg-alpha-doi}  
\bibliography{egbibsample}        

\newcommand{\etalchar}[1]{$^{#1}$}
\begin{thebibliography}{\uppercase{WGGP{\etalchar{*}}11}}

\bibitem[ADBB17]{Arulkumaran2017DeepSurvey}
\textsc{Arulkumaran K., Deisenroth M.~P., Brundage M., Bharath A.~A.}:
\newblock {Deep Reinforcement Learning: A Brief Survey}.
\newblock \emph{IEEE Signal Processing Magazine 34}, 6 (11 2017), 26--38.
\newblock \href {https://doi.org/10.1109/MSP.2017.2743240}
  {\path{doi:10.1109/MSP.2017.2743240}}.

\bibitem[BNVB13]{Bellemare2013TheAgents}
\textsc{Bellemare M.~G., Naddaf Y., Veness J., Bowling M.}:
\newblock {The arcade learning environment: An evaluation platform for general
  agents}.
\newblock \emph{Journal of Artificial Intelligence Research} (2013).

\bibitem[BOH11]{Bostock2011D3:Documents}
\textsc{Bostock M., Ogievetsky V., Heer J.}:
\newblock {D3: Data-Driven Documents}.
\newblock \emph{IEEE Trans. Visualization {\&} Comp. Graphics (Proc. InfoVis)}
  (2011).

\bibitem[BWDS19]{BeechingDeepSupercomputer}
\textsc{Beeching E., Wolf C., Dibangoye J., Simonin O.}:
\newblock {Deep Reinforcement Learning on a Budget: 3D Control and Reasoning
  Without a Supercomputer}.
\newblock \emph{arXiv preprint arXiv:1904.01806} (2019).

\bibitem[CBH19]{crosby2019animal}
\textsc{Crosby M., Beyret B., Halina M.}:
\newblock The animal-ai olympics.
\newblock \emph{Nature Machine Intelligence 1}, 5 (2019), 257.

\bibitem[CBYCT19]{cadene2019murel}
\textsc{Cadene R., Ben-Younes H., Cord M., Thome N.}:
\newblock Murel: Multimodal relational reasoning for visual question answering.
\newblock In \emph{Proceedings of the IEEE Conference on Computer Vision and
  Pattern Recognition} (2019), pp.~1989--1998.

\bibitem[CDF{\etalchar{*}}17]{Chang2017Matterport3D:Environments}
\textsc{Chang A., Dai A., Funkhouser T., Halber M., Niessner M., Savva M., Song
  S., Zeng A., Zhang Y.}:
\newblock {Matterport3D: Learning from RGB-D Data in Indoor Environments}.
\newblock \emph{International Conference on 3D Vision (3DV)} (2017).

\bibitem[CGCB14]{Chung2014EmpiricalModeling}
\textsc{Chung J., Gulcehre C., Cho K., Bengio Y.}:
\newblock {Empirical Evaluation of Gated Recurrent Neural Networks on Sequence
  Modeling}.
\newblock \emph{arXiv:1412.3555 [cs]} (12 2014).

\bibitem[CHJO16]{Carter2016ExperimentsNetworkb}
\textsc{Carter S., Ha D., Johnson I., Olah C.}:
\newblock {Experiments in Handwriting with a Neural Network}.
\newblock \emph{Distill} (2016).
\newblock \href {https://doi.org/10.23915/distill.00004}
  {\path{doi:10.23915/distill.00004}}.

\bibitem[CPMC]{CashmanRNNbow:Networks}
\textsc{Cashman D., Patterson G., Mosca A., Chang R.}:
\newblock {RNNbow: Visualizing Learning via Backpropagation Gradients in
  Recurrent Neural Networks}.
\newblock 9.

\bibitem[DDCW19]{DebardECMLPKDD2019}
\textsc{Debard Q., Dibangoye J., Canu S., Wolf C.}:
\newblock Learning 3d navigation protocols on touch interfaces with cooperative
  multi-agent reinforcement learning.
\newblock In \emph{To appear in European Conference on Machine Learning and
  Principles and Practice of Knowledge Discovery in Databases (ECML-PKDD)}
  (2019).

\bibitem[DDG{\etalchar{*}}18]{embodiedqa}
\textsc{Das A., Datta S., Gkioxari G., Lee S., Parikh D., Batra D.}:
\newblock {E}mbodied {Q}uestion {A}nswering.
\newblock In \emph{CVPR} (2018).

\bibitem[Dij59]{Dijkstra1959}
\textsc{Dijkstra E.}:
\newblock A note on two problems in connexion with graphs.
\newblock \emph{Numerische mathematik 1} (1959), 269–271.

\bibitem[DWPQ{\etalchar{*}}08]{DavidWang2008AligningRecords}
\textsc{David~Wang T., Plaisant C., Quinn A.~J., Stanchak R., Shneiderman B.,
  Murphy S.}:
\newblock \emph{{Aligning Temporal Data by Sentinel Events: Discovering
  Patterns in Electronic Health Records}}.
\newblock 2008.

\bibitem[FFM{\etalchar{*}}13]{fuchs2013evaluation}
\textsc{Fuchs J., Fischer F., Mansmann F., Bertini E., Isenberg P.}:
\newblock Evaluation of alternative glyph designs for time series data in a
  small multiple setting.
\newblock In \emph{Proceedings of the SIGCHI conference on human factors in
  computing systems} (2013), pp.~3237--3246.

\bibitem[GAW{\etalchar{*}}11]{gleicher_visual_2011}
\textsc{Gleicher M., Albers D., Walker R., Jusufi I., Hansen C.~D., Roberts
  J.~C.}:
\newblock Visual comparison for information visualization.
\newblock \emph{Information Visualization 10}, 4 (2011), 289--309.
\newblock URL: \url{http://ivi.sagepub.com/content/10/4/289.short}.

\bibitem[GKDF17]{greydanus2017visualizing}
\textsc{Greydanus S., Koul A., Dodge J., Fern A.}:
\newblock Visualizing and understanding atari agents.
\newblock \emph{arXiv preprint arXiv:1711.00138} (2017).

\bibitem[GKR{\etalchar{*}}18]{gordon2018iqa}
\textsc{Gordon D., Kembhavi A., Rastegari M., Redmon J., Fox D., Farhadi A.}:
\newblock Iqa: Visual question answering in interactive environments.
\newblock In \emph{CVPR} (2018), IEEE.

\bibitem[GSK{\etalchar{*}}17]{Greff2017LSTM:Odyssey}
\textsc{Greff K., Srivastava R.~K., Koutn{\'{i}}k J., Steunebrink B.~R.,
  Schmidhuber J.}:
\newblock {LSTM: A Search Space Odyssey}.
\newblock \emph{IEEE Transactions on Neural Networks and Learning Systems 28},
  10 (10 2017), 2222--2232.
\newblock \href {https://doi.org/10.1109/TNNLS.2016.2582924}
  {\path{doi:10.1109/TNNLS.2016.2582924}}.

\bibitem[HKPC19]{Hohman2019VisualFrontiers}
\textsc{Hohman F.~M., Kahng M., Pienta R., Chau D.~H.}:
\newblock {Visual Analytics in Deep Learning: An Interrogative Survey for the
  Next Frontiers}.
\newblock \emph{IEEE Transactions on Visualization and Computer Graphics}
  (2019).
\newblock \href {https://doi.org/10.1109/TVCG.2018.2843369}
  {\path{doi:10.1109/TVCG.2018.2843369}}.

\bibitem[HNR68]{AStar1968}
\textsc{Hart P., Nilsson N., Raphael B.}:
\newblock A formal basis for the heuristic determination of minimum cost paths.
\newblock \emph{{IEEE transactions on Systems Science and Cybernetics} 4}
  (1968), 100--107.

\bibitem[HS15]{Hausknecht2015DeepMdps}
\textsc{Hausknecht M., Stone P.}:
\newblock {Deep recurrent q-learning for partially observable mdps}.
\newblock In \emph{2015 AAAI Fall Symposium Series} (2015).

\bibitem[JBTR19]{Justesen2019DeepPlaying}
\textsc{Justesen N., Bontrager P., Togelius J., Risi S.}:
\newblock {Deep learning for video game playing}.
\newblock \emph{IEEE Transactions on Games} (2019).

\bibitem[JVW19]{Jaunet:2019}
\textsc{Jaunet T., Vuillemot R., Wolf C.}:
\newblock What if we reduce the memory of an artificial doom player?
\newblock URL: \url{https://theo-jaunet.github.io/MemoryReduction/}.

\bibitem[KAF{\etalchar{*}}08]{Keim2008VisualChallenges}
\textsc{Keim D., Andrienko G., Fekete J.-D., G{\"{o}}rg C., Kohlhammer J.,
  Melan{\c{c}}on G.}:
\newblock {Visual Analytics: Definition, Process, and Challenges}.
\newblock In \emph{Information Visualization}. Springer Berlin Heidelberg,
  2008, pp.~154--175.
\newblock \href {https://doi.org/10.1007/978-3-540-70956-5{\_}7}
  {\path{doi:10.1007/978-3-540-70956-5{\_}7}}.

\bibitem[KAL{\etalchar{*}}18]{kerpedjiev2018higlass}
\textsc{Kerpedjiev P., Abdennur N., Lekschas F., McCallum C., Dinkla K.,
  Strobelt H., Luber J.~M., Ouellette S.~B., Azhir A., Kumar N., Hwang J., Lee
  S., Alver B.~H., Pfister H., Mirny L.~A., Park P.~J., Gehlenborg N.}:
\newblock Higlass: web-based visual exploration and analysis of genome
  interaction maps.
\newblock \emph{Genome Biology 19}, 1 (Aug 2018), 125.
\newblock URL: \url{https://doi.org/10.1186/s13059-018-1486-1}, \href
  {https://doi.org/10.1186/s13059-018-1486-1}
  {\path{doi:10.1186/s13059-018-1486-1}}.

\bibitem[KCK{\etalchar{*}}18]{kwon2018retainvis}
\textsc{Kwon B.~C., Choi M.-J., Kim J.~T., Choi E., Kim Y.~B., Kwon S., Sun J.,
  Choo J.}:
\newblock Retainvis: Visual analytics with interpretable and interactive
  recurrent neural networks on electronic medical records.
\newblock \emph{IEEE transactions on visualization and computer graphics 25}, 1
  (2018), 299--309.

\bibitem[Kei00]{keim_designing_2000}
\textsc{Keim D.~A.}:
\newblock Designing {Pixel}-{Oriented} {Visualization} {Techniques}: {Theory}
  and {Applications}.
\newblock \emph{IEEE Transactions on Visualization and Computer Graphics 6}, 1
  (Jan. 2000), 59--78.
\newblock URL: \url{http://dx.doi.org/10.1109/2945.841121}, \href
  {https://doi.org/10.1109/2945.841121} {\path{doi:10.1109/2945.841121}}.

\bibitem[KJFF15]{Karpathy2015VisualizingNetworks}
\textsc{Karpathy A., Johnson J., Fei-Fei L.}:
\newblock {Visualizing and understanding recurrent networks}.
\newblock \emph{arXiv preprint arXiv:1506.02078} (2015).

\bibitem[KWR{\etalchar{*}}16]{Kempka2016ViZDoom:Learningb}
\textsc{Kempka M., Wydmuch M., Runc G., Toczek J., Ja{\'{s}}kowski W.}:
\newblock {ViZDoom: A Doom-based AI Research Platform for Visual Reinforcement
  Learning}.
\newblock In \emph{IEEE Conference on Computational Intelligence and Games} (9
  2016), IEEE, pp.~341--348.

\bibitem[LB04]{Light2004TheGraphics}
\textsc{Light A., Bartlein P.~J.}:
\newblock {The end of the rainbow? Color schemes for improved data graphics}.
\newblock \emph{Eos, Transactions American Geophysical Union 85}, 40 (2004),
  385--391.

\bibitem[LC17]{Lample2017PlayingLearning}
\textsc{Lample G., Chaplot D.~S.}:
\newblock {Playing FPS games with deep reinforcement learning}.
\newblock In \emph{Thirty-First AAAI Conference on Artificial Intelligence}
  (2017).

\bibitem[LCM{\etalchar{*}}18]{lehman2018surprising}
\textsc{Lehman J., Clune J., Misevic D., Adami C., Altenberg L., Beaulieu J.,
  Bentley P.~J., Bernard S., Beslon G., Bryson D.~M., et~al.}:
\newblock The surprising creativity of digital evolution: A collection of
  anecdotes from the evolutionary computation and artificial life research
  communities.
\newblock \emph{arXiv preprint arXiv:1803.03453} (2018).

\bibitem[LFDA16]{Levine2016End-to-endPoliciesb}
\textsc{Levine S., Finn C., Darrell T., Abbeel P.}:
\newblock {End-to-end training of deep visuomotor policies}.
\newblock \emph{The Journal of Machine Learning Research 17}, 1 (2016),
  1334--1373.

\bibitem[LGS{\etalchar{*}}14]{lex2014upset}
\textsc{Lex A., Gehlenborg N., Strobelt H., Vuillemot R., Pfister H.}:
\newblock Upset: visualization of intersecting sets.
\newblock \emph{IEEE transactions on visualization and computer graphics 20},
  12 (2014), 1983--1992.

\bibitem[Lip16]{Lipton2016TheInterpretability}
\textsc{Lipton Z.~C.}:
\newblock {The mythos of model interpretability}.
\newblock \emph{arXiv preprint arXiv:1606.03490} (2016).

\bibitem[MCZ{\etalchar{*}}17]{Ming2017UnderstandingNetworks}
\textsc{Ming Y., Cao S., Zhang R., Li Z., Chen Y., Song Y., Qu H.}:
\newblock {Understanding Hidden Memories of Recurrent Neural Networks}.
\newblock \emph{arXiv:1710.10777 [cs]} (10 2017).

\bibitem[MKS{\etalchar{*}}13]{Mnih2013PlayingLearning}
\textsc{Mnih V., Kavukcuoglu K., Silver D., Graves A., Antonoglou I., Wierstra
  D., Riedmiller M.}:
\newblock {Playing Atari with Deep Reinforcement Learning}.
\newblock \emph{arXiv:1312.5602 [cs]} (12 2013).

\bibitem[MKS{\etalchar{*}}15]{mnih2015humanlevel}
\textsc{Mnih V., Kavukcuoglu K., Silver D., Rusu A.~A., Veness J., Bellemare
  M.~G., Graves A., Riedmiller M., Fidjeland A.~K., Ostrovski G., Petersen S.,
  Beattie C., Sadik A., Antonoglou I., King H., Kumaran D., Wierstra D., Legg
  S., Hassabis D.}:
\newblock Human-level control through deep reinforcement learning.
\newblock \emph{Nature 518}, 7540 (2015).

\bibitem[MPBM{\etalchar{*}}16]{Mnih2016AsynchronousLearning}
\textsc{Mnih V., Puigdom{\`{e}}nech~Badia A., Mirza M., Graves A., Harley T.,
  Lillicrap T.~P., Silver D., Kavukcuoglu K.}:
\newblock \emph{{Asynchronous Methods for Deep Reinforcement Learning}}.
\newblock Tech. rep., 2016.

\bibitem[MPV{\etalchar{*}}16]{Mirowski2016LearningEnvironments}
\textsc{Mirowski P., Pascanu R., Viola F., Soyer H., Ballard A.~J., Banino A.,
  Denil M., Goroshin R., Sifre L., Kavukcuoglu K., {others}}:
\newblock {Learning to navigate in complex environments}.
\newblock \emph{arXiv preprint arXiv:1611.03673} (2016).

\bibitem[OCSL16]{Oh2016ControlMinecraftb}
\textsc{Oh J., Chockalingam V., Singh S., Lee H.}:
\newblock {Control of Memory, Active Perception, and Action in Minecraft}.

\bibitem[OSJ{\etalchar{*}}18]{olah2018the}
\textsc{Olah C., Satyanarayan A., Johnson I., Carter S., Schubert L., Ye K.,
  Mordvintsev A.}:
\newblock The building blocks of interpretability.
\newblock \emph{Distill} (2018).
\newblock https://distill.pub/2018/building-blocks.
\newblock \href {https://doi.org/10.23915/distill.00010}
  {\path{doi:10.23915/distill.00010}}.

\bibitem[PGC{\etalchar{*}}17]{Paszke2017AutomaticPyTorch}
\textsc{Paszke A., Gross S., Chintala S., Chanan G., Yang E., DeVito Z., Lin
  Z., Desmaison A., Antiga L., Lerer A.}:
\newblock {Automatic differentiation in PyTorch}.

\bibitem[PS18]{ParisottoNeuralMap2018}
\textsc{Parisotto E., Salakhutdinov R.}:
\newblock Neural map: Structured memory for deep reinforcement learning.
\newblock \emph{ICLR} (2018).

\bibitem[RSG16]{ribeiro2016should}
\textsc{Ribeiro M.~T., Singh S., Guestrin C.}:
\newblock Why should i trust you?: Explaining the predictions of any
  classifier.
\newblock In \emph{Proceedings of the 22nd ACM SIGKDD international conference
  on knowledge discovery and data mining} (2016), ACM, pp.~1135--1144.

\bibitem[SB18]{Sutton2018ReinforcementIntroduction}
\textsc{Sutton R.~S., Barto A.~G.}:
\newblock \emph{{Reinforcement learning: An introduction}}.
\newblock 2018.

\bibitem[SCD{\etalchar{*}}17]{selvaraju2017grad}
\textsc{Selvaraju R.~R., Cogswell M., Das A., Vedantam R., Parikh D., Batra
  D.}:
\newblock Grad-cam: Visual explanations from deep networks via gradient-based
  localization.
\newblock In \emph{Proceedings of the IEEE international conference on computer
  vision} (2017), pp.~618--626.

\bibitem[SDBR14]{Springenberg2014StrivingNetc}
\textsc{Springenberg J.~T., Dosovitskiy A., Brox T., Riedmiller M.}:
\newblock {Striving for Simplicity: The All Convolutional Net}.

\bibitem[SGPR17]{Strobelt2017LSTMVis:Networks}
\textsc{Strobelt H., Gehrmann S., Pfister H., Rush A.~M.}:
\newblock {Lstmvis: A tool for visual analysis of hidden state dynamics in
  recurrent neural networks}.
\newblock \emph{IEEE transactions on visualization and computer graphics 24}, 1
  (2017), 667--676.

\bibitem[SKM{\etalchar{*}}19]{Savva2019Habitat:Research}
\textsc{Savva M., Kadian A., Maksymets O., Zhao Y., Wijmans E., Jain B., Straub
  J., Liu J., Koltun V., Malik J., Parikh D., Batra D.}:
\newblock {Habitat: A Platform for Embodied AI Research}.

\bibitem[SSS{\etalchar{*}}17]{Silver2017MasteringKnowledge}
\textsc{Silver D., Schrittwieser J., Simonyan K., Antonoglou I., Huang A., Guez
  A., Hubert T., Baker L., Lai M., Bolton A., Chen Y., Lillicrap T., Hui F.,
  Sifre L., van~den Driessche G., Graepel T., Hassabis D.}:
\newblock {Mastering the game of Go without human knowledge}.
\newblock \emph{Nature 550}, 7676 (10 2017), 354--359.
\newblock \href {https://doi.org/10.1038/nature24270}
  {\path{doi:10.1038/nature24270}}.

\bibitem[STST18]{Salomon2018HumanAutomata}
\textsc{Salom{\'{o}}n S., T{\^{i}}rn{\u{a}}uc{\u{a}} C., Salom{\'{o}}n S.,
  T{\^{i}}rn{\u{a}}uc{\u{a}} C.}:
\newblock {Human Activity Recognition through Weighted Finite Automata}.
\newblock \emph{Proceedings 2}, 19 (10 2018), 1263.
\newblock \href {https://doi.org/10.3390/proceedings2191263}
  {\path{doi:10.3390/proceedings2191263}}.

\bibitem[TGF09]{tang2009exploring}
\textsc{Tang A., Greenberg S., Fels S.}:
\newblock Exploring video streams using slit-tear visualizations.
\newblock In \emph{CHI'09 Extended Abstracts on Human Factors in Computing
  Systems} (2009), ACM, pp.~3509--3510.

\bibitem[TH12]{tieleman2012lecture}
\textsc{Tieleman T., Hinton G.}:
\newblock Lecture 6.5-rmsprop: Divide the gradient by a running average of its
  recent magnitude.
\newblock \emph{COURSERA: Neural networks for machine learning 4}, 2 (2012),
  26--31.

\bibitem[VDMH08]{VanDerMaaten2008VisualizingT-SNE}
\textsc{Van Der~Maaten L., Hinton G.}:
\newblock {Visualizing Data using t-SNE}.
\newblock \emph{Journal of Machine Learning Research 9} (2008), 2579--2605.

\bibitem[WGGP{\etalchar{*}}11]{Wongsuphasawat2011LifeFlow:Sequences}
\textsc{Wongsuphasawat K., Guerra~G{\'{o}}mez J.~A., Plaisant C., Wang T.~D.,
  Shneiderman B., Taieb-Maimon M.}:
\newblock \emph{{LifeFlow: Visualizing an Overview of Event Sequences}}.
\newblock 2011.

\bibitem[WGSY18]{Wang2018Dqnviz:Q-networks}
\textsc{Wang J., Gou L., Shen H.-W., Yang H.}:
\newblock {Dqnviz: A visual analytics approach to understand deep q-networks}.
\newblock \emph{IEEE transactions on visualization and computer graphics 25}, 1
  (2018), 288--298.

\bibitem[Yam97]{YamauchiFrontiers1997}
\textsc{Yamauchi B.}:
\newblock {A frontier-based approach for autonomous exploration}.
\newblock In \emph{{Symposium on Computational Intelligence in Robotics and
  Automation}} (1997).

\bibitem[ZBZM16]{Zahavy2016GrayingDqnsb}
\textsc{Zahavy T., Ben-Zrihem N., Mannor S.}:
\newblock {Graying the black box: Understanding dqns}.
\newblock In \emph{International Conference on Machine Learning} (2016),
  pp.~1899--1908.

\bibitem[ZKL{\etalchar{*}}16]{zhou2016learning}
\textsc{Zhou B., Khosla A., Lapedriza A., Oliva A., Torralba A.}:
\newblock Learning deep features for discriminative localization.
\newblock In \emph{Proceedings of the IEEE conference on computer vision and
  pattern recognition} (2016), pp.~2921--2929.

\bibitem[ZMK{\etalchar{*}}17]{Zhu2017Target-drivenLearning}
\textsc{Zhu Y., Mottaghi R., Kolve E., Lim J.~J., Gupta A., Fei-Fei L., Farhadi
  A.}:
\newblock {Target-driven visual navigation in indoor scenes using deep
  reinforcement learning}.
\newblock In \emph{2017 IEEE international conference on robotics and
  automation (ICRA)} (2017), pp.~3357--3364.

\end{thebibliography}



\end{document}